%% file: main.tex
\documentclass[letterpaper, 10pt, conference]{IEEEtran}

\usepackage{cite}
\usepackage{float}
\usepackage{amsmath}
\usepackage{amssymb}
\usepackage{amsfonts}
\usepackage{array}
\usepackage{tabularx}
\usepackage{breqn}
\usepackage{url}
\usepackage[colorlinks=true, allcolors=black]{hyperref}
\usepackage{csvsimple}
\usepackage{caption}
\usepackage{tikz}
\usepackage{tikzscale}
\usepackage{svg}
\usetikzlibrary{arrows}
\usepackage{multirow}
\usepackage{algorithm}
\usepackage{algpseudocode}
\usepackage{algcompatible}
\usepackage[english]{babel}
\usepackage{amsthm}
\newtheorem{theorem}{Theorem}
\usepackage{graphicx}
\usepackage{subcaption}
\usepackage{scalefnt}
\usepackage{capt-of}
\usepackage{booktabs}
\usepackage{comment}
\usepackage{multimedia}
\usepackage{lastpage}
\usepackage{fancyhdr}
\usepackage{siunitx}
\usepackage{booktabs}
\usepackage{csquotes}
\usepackage{makecell}

\usepackage{xcolor}
\usepackage{soul}
\sethlcolor{yellow} 

\input{utils/functions}

\begin{document}



\title{\vspace*{0.3in} Beyond the Snowflakes: Disentangling Weather Effects from Confounding Meta-Factors in LiDAR Object Detection}

\author{
    Yiming~Yang$^*$,
    Jeremy~P. Bos$^\ddagger$
    \thanks{$^*$$^\dagger$$^\ddagger$ Y. Yang and J. Bos are with the Department of Electrical and Computer Engineering, Michigan Technological University, Houghton,
    MI, USA.}
    \thanks{$^*$yimingya@mtu.edu, $^\ddagger$jpbos@mtu.edu}
}

\maketitle

\thispagestyle{empty}

\begin{abstract}
Light Detection and Ranging (LiDAR) is critical for autonomous perception, yet its performance is known to degrade in adverse weather. While existing literature models snow as atmospheric noise that degrades performance, our granular evaluation on the WADS-3D and CADC datasets reveals a counter-intuitive phenomenon: detection accuracy consistently improves as snow intensity increases. Through ablation study and meta-information analysis,  we reconcile this by confirming that while falling snow acts as noise, its negative impact is outweighed by a dominant, concurrent factor: significantly increased object point count. We hypothesize that this object point count boost acts as a geometric enhancement, driven by snow accumulation increasing vehicle surface area and a shift from specular to diffuse reflectance. Our findings demonstrate that evaluating perception systems using broad weather-severity bins without accounting for correlated object-level data can lead to misleading conclusions about detector robustness. We conclude that accurate evaluation requires decoupling atmospheric noise from surface accumulation effects and outline future research to validate these physical models. The WADS-3D dataset is available at: \url{https://digitalcommons.mtu.edu/wads-3d/}.

\end{abstract}

\begin{IEEEkeywords}
Lidar, Snowfall, Object Detection, Autonomous vehicles, Adverse weather
\end{IEEEkeywords}

\IEEEpeerreviewmaketitle



\section{Introduction}
\label{sec:introduction}
\input{sections/introduction.tex}

\section{Related Works}
\label{sec:relatedworks}
\input{sections/relatedworks.tex}

\section{Methods}
\label{sec:methods}
\input{sections/methods.tex} 

\section{Experiments}
\label{sec:experiments}
\input{sections/experiments.tex}

\section{Conclusion and future work}
\label{sec:conclusion}
\input{sections/conclusion.tex}


\bibliographystyle{IEEEtran}
\bibliography{bibliography/references}

\ifCLASSOPTIONcaptionsoff
  \newpage
\fi

\end{document}

%% file: utils/functions.tex
\newtheorem{lemma}[theorem]{Lemma}

\newcolumntype{C}{>{\centering\arraybackslash}X}



%% file: sections/introduction.tex
\IEEEPARstart{L}{idar} sensors are a cornerstone of perception for robotics and autonomous vehicles due to their high-resolution 3D measurement capabilities. In these systems, LiDAR enables critical tasks such as localization \cite{9304577}, obstacle avoidance \cite{9906811}, and object detection \cite{9181591}, providing precise depth information where vision-only systems may struggle.

However, adverse weather conditions including rain, fog, and snow degrade LiDAR performance by introducing noise, occlusion, and false returns \cite{10186539}. Snowfall, in particular, presents unique challenges: snowflakes act as scatterers that obscure objects and create transient noise points. For autonomous systems, these errors can propagate to downstream modules \cite{10439634, courcelle2023importance, 9981724}, potentially leading to safety-critical failures.

Despite these risks, many perception algorithms are developed under ideal weather conditions, leaving a significant reliability gap in real-world, dynamic environments. While recent studies have begun investigating LiDAR performance in fog \cite{Hahner_2021_ICCV}, rain \cite{10889253}, and snow \cite{Hahner_2022_CVPR}, snowfall research remains comparatively limited. Investigating snow is difficult due to its complex spatiotemporal characteristics \cite{li2023snow}, the lack of controlled test environments \cite{tang2025hardsnowpaireddomain}, and the high cost of data acquisition. Consequently, many studies rely on theoretical scattering models \cite{10920127} or qualitative observations, leaving a gap in systematic, quantitative analysis across varying snowfall intensities.

To bridge these gaps, we introduce the WADS 3D Object Detection Dataset (WADS-3D), an extension of the Winter Adverse Driving Dataset (WADS) \cite{kurup2023winter}. Leveraging multi-year data from Michigan’s Upper Peninsula, WADS captures snowfall intensities significantly higher than those typically reported in existing literature \cite{bos2023assessing}, providing a unique benchmark for the most severe winter conditions. To ensure our findings generalize across different environments and sensor specifications, we further evaluate our results against the established Canadian Adverse Driving Conditions Dataset (CADC) \cite{pitropov2021canadian}. By utilizing WADS-3D for extreme-case analysis alongside CADC for broader validation, we present a comprehensive evaluation of 3D object detection performance across a wide spectrum of winter severity. Our analysis reveals a counter-intuitive key finding: detection performance can improve with heavier snowfall. To explain this, we conduct
an ablation study, error matrix analysis, and a three-stage meta information analysis,
which demonstrates that the improvement is not a direct result
of the falling snow, but rather an artifact of confounding
variables. The model’s performance is mainly governed by
the object complexity rather than environmental complexity.

\begin{figure}[htp]
\centering\includegraphics[width=8.8cm]{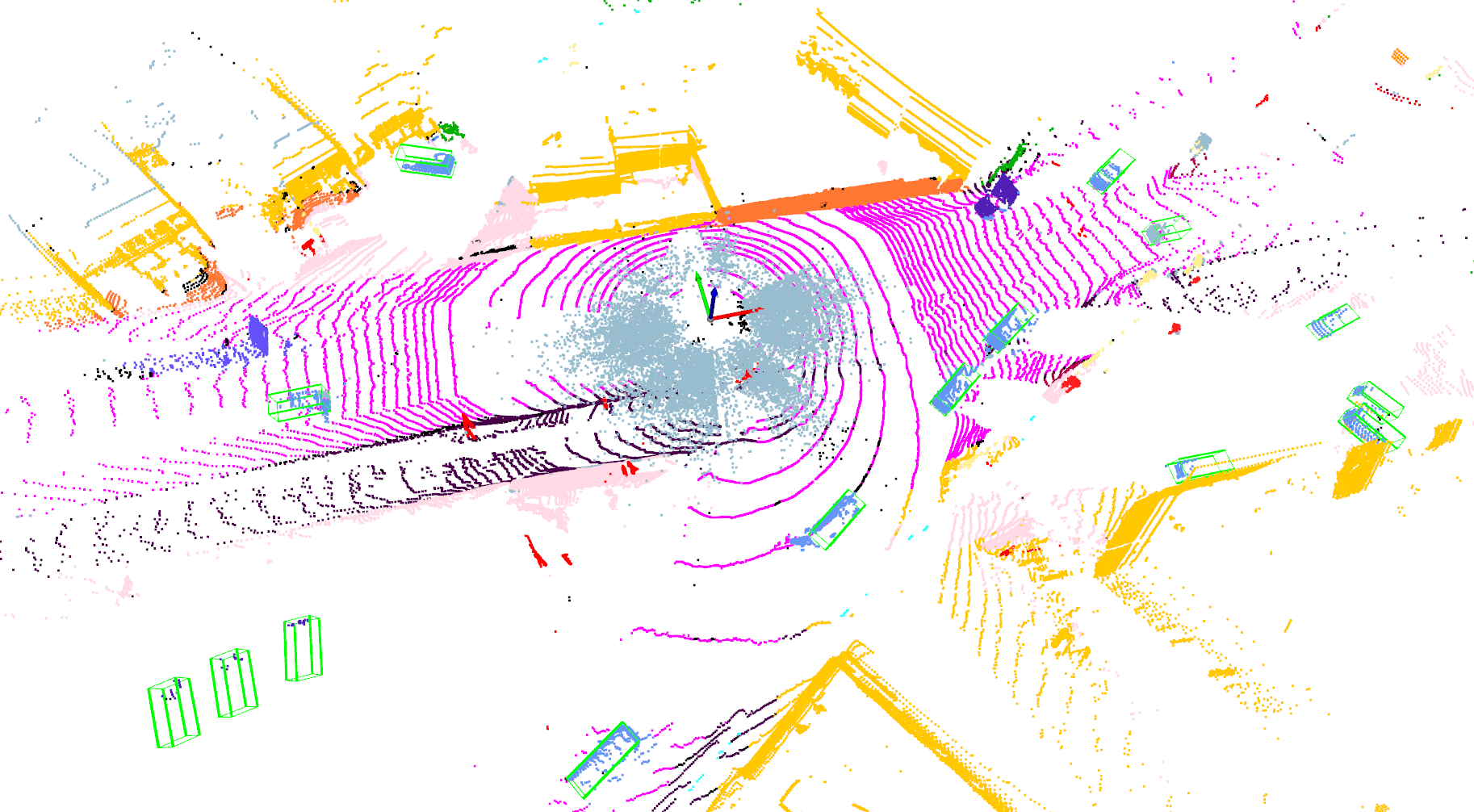}\caption{A labeled frame from WADS-3D, with points colored according to the SemanticKITTI style. Cars are indicated by green bounding boxes.}\label{fig:frame_from_wads}\vspace{-3mm}
\end{figure}

%% file: sections/relatedworks.tex
\subsection{LiDAR-based 3D Object Detection}

LiDAR-based 3D object detection identifies and localizes objects within sparse and irregular point clouds. Traditional approaches utilized geometry-based clustering (e.g., Euclidean \cite{rusu2010semantic} or Supervoxel \cite{papon2013voxel}) alongside hand-crafted features and simple classifiers \cite{zhao2021technical}. Modern methods are dominated by deep neural networks, categorized by their data representation: Voxel-based methods \cite{chen2023voxelnextfullysparsevoxelnet} discretize space into 3D grids for efficient sparse convolutions (e.g., SECOND \cite{yan2018second}, Voxel R-CNN \cite{deng2021voxel}); Pillar-based methods \cite{lang2019pointpillars} simplify this into 2D vertical columns for faster processing; Point-based architectures \cite{shi2019pointrcnn} operate directly on raw points; and Hybrid models \cite{shi2020pv, shi2020points} integrate both voxel and point features. While sensor fusion with cameras \cite{Bai_2022_CVPR}, radar \cite{10611436} or 4D-radar \cite{Chae_2024_CVPR} often achieves state-of-the-art results, this study focuses exclusively on LiDAR-only methods to isolate performance degradation specifically caused by snowy conditions.

\subsection{Impact of Snow on LiDAR Perception}  

Snowfall disrupts LiDAR perception through three primary mechanisms: backscatter noise (false positives), occlusion \cite{kurup2021dsor}, and signal attenuation \cite{10920127}. Studies utilizing physically-based simulations, such as Hahner et al. \cite{Hahner_2022_CVPR}, typically report a monotonic degradation in detection performance driven by these factors. However, these simulations primarily model airborne snow, neglecting the geometric effects of snow accumulation on object surfaces and the surrounding environment. In real-world evaluations, Sun et al. \cite{10556562} quantified degradation on the MSP dataset, reporting vehicle detection drops of up to 50\% under heavy snow. Yet, this analysis aggregated multiple heterogeneous datasets (Boreas\cite{burnett2023boreas}, CADC\cite{pitropov2021canadian}, RADIATE\cite{sheeny2021radiate}, and WADS\cite{kurup2023winter}) collected with varying LiDAR sensors. Merging such diverse data introduces sensor-specific domain shifts that are difficult to eliminate entirely, potentially confounding weather effects with sensor discrepancies. Complementing this, the CADC+ dataset \cite{tang2025hardsnowpaireddomain} demonstrated that snow introduces both aleatoric and epistemic uncertainties, effectively acting as a distinct data domain.

\subsection{Object- and Scene-Level Complexity Metrics}

Complexity is a decisive factor in perception performance, yet metrics vary across benchmarks. At the object level, KITTI \cite{geiger2012we} and A*3D \cite{9197385} use camera-centric properties like pixel height and occlusion. Conversely, the Waymo Open Dataset \cite{sun2020scalability} employs LiDAR-centric metrics (e.g., object point counts), while nuScenes \cite{caesar2020nuscenes} relies on post-hoc error analysis (NDS) correlated against visibility and range. At the scene level, complexity is defined by task-based or geometric factors. SemanticKITTI \cite{behley2021towards} uses manual annotation effort as a proxy for complexity, while Abohassan et al. \cite{s24020452} quantify the computational load imposed on LiDAR systems by dense infrastructure. Geometric approaches include surface smoothness derivatives \cite{chen2017geometry}, semantic occupancy grids (M-SOG) \cite{NEURIPS2024_3dbb8b6b}, and information-theoretic measures such as LiDAR point cloud entropy \cite{liu2022novel} or aerial image delentropy \cite{beche2025claravidholisticscenereconstruction}.

\subsection{Analyzing the Influence of Meta-Information}

The black-box nature of deep learning detectors poses challenges for safety certification. To improve transparency, researchers analyze how specific meta-information influences performance. While Zendel et al. \cite{Zendel_2023_CVPR} and Linnhoff et al. \cite{linnhoff2021towards} identified hazard chains for LiDAR and cameras, but their work lacked quantitative deep-learning analysis. Ponn et al. \cite{ponn2020performance} introduced a multivariate approach using Random Forest (RF) and SHapley Additive exPlanations (SHAP) to analyze 2D camera-based detectors. Recently, Kuznietsov \cite{kuznietsov2025methodologyanalysisinfluencingfactors} applied Kendall’s Tau and SHAP to rank influencing factors for LiDAR-based pedestrian detection on nuScenes. By analyzing both False Negative (FN) and False Positive (FP) errors, they highlighted the importance of accounting for multivariate dependencies to improve analytical reliability.

\subsection{Research Gap}

Current literature on snowy LiDAR perception is limited by reliance on coarse weather categories and synthetic models that typically simulate atmospheric noise while ignoring surface accumulation. Furthermore, empirical studies often merge heterogeneous datasets collected with varying sensors, introducing domain shifts that obscure genuine physical trends. Consequently, simply measuring accuracy drops fails to distinguish whether performance changes stem from atmospheric occlusion, object-level geometric changes, or sensor inconsistencies. To address these gaps, this work provides the following contributions: (1) we introduce WADS-3D, a dataset featuring extreme snowfall intensities for LiDAR perception; (2) we conduct independent benchmarks on WADS-3D and CADC to validate performances across distinct environments and sensor configurations without domain-shift artifacts; and (3) we employ meta-information analysis to reveal that object complexity primarily drives detector performance in heavy snow.

%% file: sections/methods.tex
\subsection{Dataset}

 WADS-3D (Fig. \ref{fig:frame_from_wads}) is derived from the Winter Adverse Driving Dataset \cite{kurup2023winter}. Following the KITTI format, it comprises 4,100 labeled frames generated by rotating 2,050 unique 360-degree scans to satisfy forward-view requirements. It contains 30,104 labeled cars; other classes were excluded due to sparsity in snowy scenes. To ensure cross-sensor validity, we also utilize the CADC dataset \cite{pitropov2021canadian}, which provides 7,000 LiDAR frames captured with a different sensor suite.

\begin{figure}[htp]
    \centering
    \includegraphics[width=8.8cm]{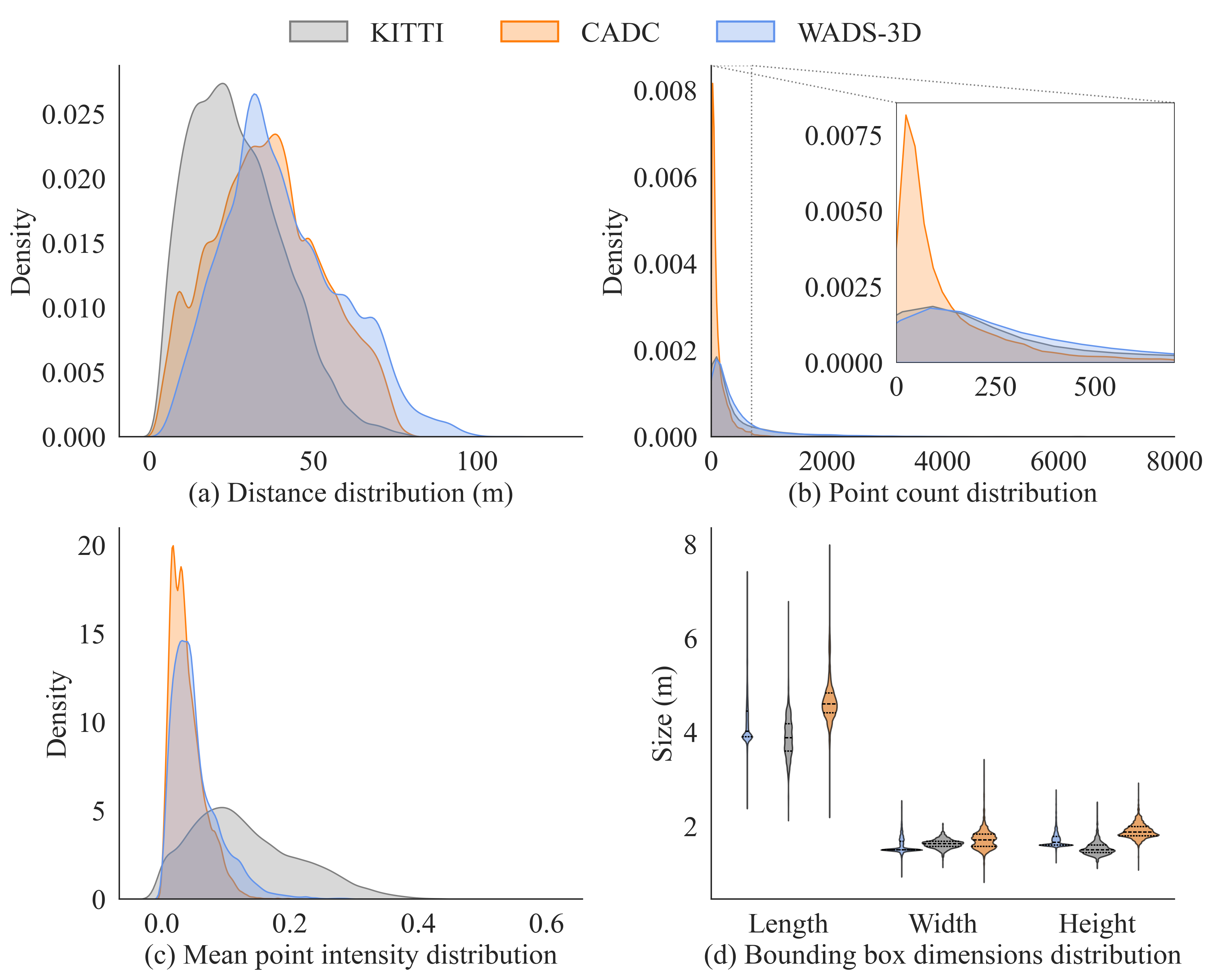}
    \caption{Key object attributes comparison among the KITTI, CADC, and WADS-3D datasets. (a) Car distance distribution, showing WADS-3D and CADC observe vehicles at greater average distances. (b) LiDAR points-per-car distribution, highlighting a higher point count in WADS-3D, followed by KITTI and CADC; the lower count in CADC is attributable to its 32-channel lidar. (c) Mean LiDAR intensity distribution, indicating lower intensity values for the snowy CADC and WADS-3D datasets. (d) Car dimensions distribution.}
    \label{fig:datasets_comparison}
    \vspace{-3mm}
\end{figure}

As shown in Fig. \ref{fig:datasets_comparison}, object-level distributions differ significantly across clear-weather KITTI, snowy CADC, and snowy WADS-3D. Despite hardware differences (Table \ref{tab:sensor_comparison}), WADS-3D vehicles exhibit a higher average point count per bounding box. These vehicles are generally observed at greater distances with lower mean intensity. Physical car dimensions also vary across datasets, with length and width following the order CADC \textgreater KITTI \textgreater WADS-3D, while height follows CADC \textgreater WADS-3D \textgreater KITTI.

\begin{table}[t!]
\centering
\caption{Comparison of LiDAR Sensor Specifications in KITTI, CADC, and WADS-3D datasets}
\label{tab:sensor_comparison}
\resizebox{\columnwidth}{!}{%
\begin{tabular}{lccc}
\toprule
\textbf{Parameter} & \textbf{Velodyne HDL-64E} & \textbf{Ouster OS1-64} & \textbf{Velodyne VLP-32C}\\
\textit{} & \textit{(KITTI Dataset)} & \textit{(WADS-3D Dataset)} & \textit{(CADC Dataset)}\\
\midrule
\makecell[l]{Channels}  & 64  & 64   & 32 

\\

\addlinespace
\makecell[l]{Vertical \\ FOV}  & \makecell{26.8\textdegree{} \\ ( -24.8\textdegree{} - +2\textdegree{})} & \makecell{32.2\textdegree{} \\ (\textpm16.6\textdegree{})} & \makecell{40\textdegree{} \\ (-25\textdegree{} - +15\textdegree{})} 

\\

\addlinespace
\makecell[l]{Vertical \\ Resolution} & \makecell{Non-uniform \\ (Avg. 0.4\textdegree{})} & \makecell{Uniform \\ (Avg. 0.52\textdegree{})} & \makecell{Non-uniform \\ (Avg. 0.33\textdegree{})} 

\\

\addlinespace 
\makecell[l]{Horizontal \\ Resolution}   & \makecell{$\sim$ 0.17\textdegree{} \\ (@10 Hz)} & \makecell{$\sim$ 0.17\textdegree{} \\ (@2048 mode)} & \makecell{$\sim$ 0.2\textdegree{} \\ (@10 Hz)}

\\

\addlinespace
\makecell[l]{Configurable \\ Modes} & \makecell{Rotation Rate \\ (5 - 20 Hz)} & \makecell{Points/Rev. \\ (512, 1024, 2048)} & \makecell{Rotation Rate \\ (5 - 20 Hz)}

\\

\addlinespace
Rotation Rate    & 10 Hz  & 10 Hz / 20 Hz  & 10 Hz / 5-20 Hz  

\\

\addlinespace
Raw Intensity     & 0-255 & 0-255 & 0-255 

\\

\addlinespace
\makecell{Normalized \\ Intensity} & 0.0-1.0 & 0.0-1.0 &  0.0-1.0 \\
\bottomrule
\end{tabular}%
}
\vspace{-3mm}
\end{table}

\subsection{Snowfall Intensity Classification}

To enable stratified analysis, we bin frames by snowfall intensity using a two-stage hybrid filter to quantify falling snow points. We first apply the Dynamic Statistical Outlier Removal (DSOR) filter \cite{kurup2021dsor} to identify snow based on spatial distribution, followed by the Low-Intensity Outlier Removal (LIOR) filter \cite{park2020fast} to isolate low-intensity returns characteristic of snowflakes. The filtering process reveals higher overall severity in WADS-3D, with snow points ranging from 764--19,386 (0.69\%--28.34\%, avg. 8.97\%) of the point cloud, compared to 142--4,447 points (0.66\%--21.38\%, avg. 6.71\%) in CADC. Snowfall intensity is categorized by these absolute counts within the forward-facing view (Fig.~\ref{fig:frame_from_wads}), ensuring a granular quantification of weather severity across both platforms. Rather than fixed intervals, we use quantile binning to ensure each snowfall intensity bin contains a comparable number of ground-truth \enquote{Car} objects, maintaining statistical consistency across varying weather severities.

\subsection{Evaluation Metircs}

To quantitatively assess detection performance, we utilize the standard evaluation protocol from the KITTI 3D object detection benchmark. Our primary metric is Average Precision (AP), which summarizes the precision-recall (PR) curve to provide a comprehensive measure of detector quality. For the ablation study, we leverage AP to maintain consistency with our main results and allow for an explicit performance breakdown. 

\subsection{Implementation Details}

\begin{figure*}[t] 
  \centering
  \includegraphics[width=0.99\textwidth]{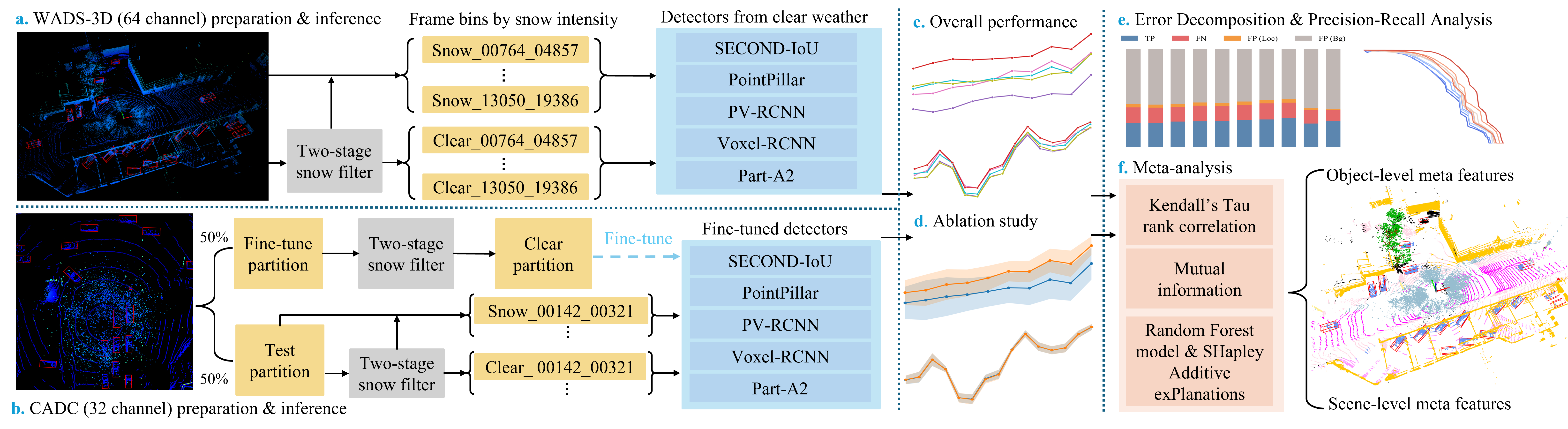} 
  \caption{Lidar-only 3D object detection pipeline, including the following sequential stages: (a) WADS-3D preparation: We binned the lidar frames using a two-stage snow filter, generating snow-filtered bins and original snow bins. We then directly applied five KITTI-trained detectors (which use a 64-channel LiDAR setting) to these groups. (b) CADC fine-tuning: We split the CADC dataset 50:50 for fine-tuning and testing, respectively. We first fine-tuned the models for 32-channel capability and then used the fine-tuned detectors for evaluation. (c) Overall performance: We aggregated and evaluated the detection results inferred by the different detectors on both datasets. (d) Ablation study: We used ablation studies to analyze the influence of varying falling snow intensities. (e) Error decomposition and precision-recall analysis. (f) A three-stage meta-information analysis is carried out on both datasets to identify key influencing meta-features at both the object and scene levels.}
  \label{fig:evaluation_pipeline}
  \vspace{-3mm}
\end{figure*}

We evaluate five detectors representing a diverse cross-section of 3D perception architectures (Table \ref{table:detectors}) from the OpenPCDet \cite{od2020openpcdet} and MMDetection3D\cite{mmdet3d2020} frameworks. By spanning voxel, pillar, point-based, and hybrid paradigms, we verify that our findings are not architectural artifacts but represent a generalized phenomenon across different discretization strategies and detection stages. We then employ two distinct evaluation strategies (Fig. \ref{fig:evaluation_pipeline}) tailored to the sensor characteristics of our target datasets: 

WADS-3D: Models were used in their pre-trained state (trained on clear-weather KITTI). This assesses how well models generalize to snow given similar 64-channel LiDAR hardware.

CADC: Due to the 32-channel sensor difference, we used a 50:50 train/eval split for fine-tuning. Training data was pre-processed with our two-stage snow filter to provide a \enquote{clean} baseline, while evaluation was performed on original, unfiltered snowy frames.

\begin{table}[t]
\caption{Architectural Classification of the Evaluated 3D Detectors}
\label{table:detectors}
\centering
\begin{tabularx}{0.95\columnwidth}{@{}lXXX@{}} 
\toprule
\textbf{Model} & \textbf{Architecture} & \textbf{Stages} & \textbf{Paradigm} \\ \midrule
SECOND (IoU)\cite{yan2018second} & Voxel-based & One-stage & IoU-aware \\
Voxel R-CNN \cite{deng2021voxel} & Voxel-based & Two-stage & Sparse Conv \\
PointPillars \cite{lang2019pointpillars} & Pillar-based & One-stage & 2D Projection \\
PV-RCNN \cite{shi2020pv} & Hybrid & Two-stage & Voxel-Point \\
Part-A2 \cite{shi2020points} & Point-based & Two-stage & Part-aware \\ \bottomrule
\end{tabularx}
\vspace{-3mm}
\end{table}

\subsection{Analysis Methodology}

To disentangle snowfall's impact from confounding variables, we employ a multi-stage framework. First, an ablation study compares detector performance on raw versus de-noised data to isolate the direct effect of airborne particles. Second, an error matrix analysis determines how specific error types fluctuate with snowfall intensity. Finally, a meta-information Analysis \cite{kuznietsov2025methodologyanalysisinfluencingfactors} to rank influencing factors using Kendall’s Tau for univariate trends, Mutual Information for non-linear relationships, and RF with SHAP analysis for multivariate dependencies.

The RF models map meta-information to detection outcomes (FN, TP), optimized via grid search. We utilize conditional sampling during SHAP computation to prevent extrapolation issues with correlated features. Meta-information features are divided into two categories: 1) Object-level features including object point count, object distance, object dimensions (length, width, height), and object point intensity; and 2) Scene-level features including traffic density, snowfall intensity, and local complexity metrics such as spatial entropy \cite{liu2022novel}, voxel occupancy \cite{NEURIPS2024_3dbb8b6b}, and geometric complexity (derived from normal vector variance) \cite{chen2017geometry}.

%% file: sections/experiments.tex
\subsection{Overall Detection Performance}

\subsubsection{Results on WADS-3D} Evaluation of the WADS-3D dataset reveals an unexpected trend: a clear positive correlation between snowfall intensity and AP across all architectures in Fig.~\ref{fig:ap_wads3d_cadc}(a). To isolate the impact of falling snow, we conducted ablation study and re-evaluated detector performance after applying our snow filter to remove atmospheric noise. As shown in Fig.~\ref{fig:ap_wads3d_cadc}(b), filtering improves absolute scores across all intensities. Crucially, however, the positive performance trend remains unchanged. This suggests that falling snow is not the driver of the improvement, but rather a proxy for a confounding variable.

\subsubsection{Results on CADC} 
\begin{figure}[H] \centering \includegraphics[width=8.5cm]{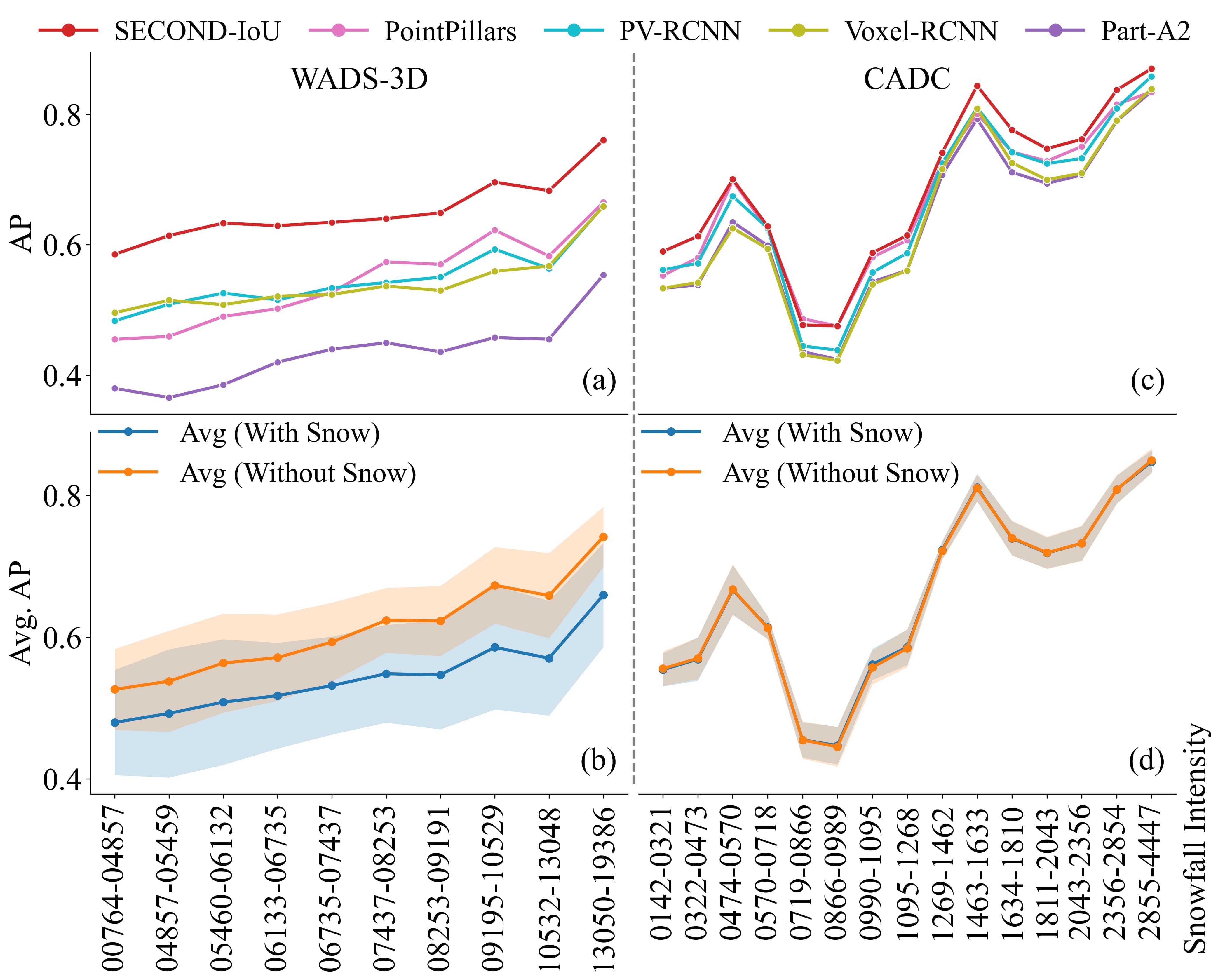} \caption{Impact of snowfall intensity on object detection performance in WADS-3D (left) and CADC (right). Top row (a,c): AP of five individual models under snowy conditions across increasing intensity. Bottom row (b,d): Ablation study, comparing the mean AP of all models with snow (blue) vs. without snow (orange) across the same intensity bins. Shaded regions indicate standard deviation.}\label{fig:ap_wads3d_cadc}\vspace{-3mm}
\end{figure}

The CADC evaluation confirms this positive trend in Fig.~\ref{fig:ap_wads3d_cadc}(c). Unlike the stable increase in WADS-3D, CADC exhibits higher volatility across snow bins. Snow filtering doesn't yields higher performance but preserves the underlying positive trend in Fig.~\ref{fig:ap_wads3d_cadc}(d), reinforcing the hypothesis that confounding variables drive the observed behavior.

\subsubsection{Impact of Explicit Snow Filtering}

A divergence appears when evaluating the impact of snow filtering across two datasets. As shown in Fig.~\ref{fig:ap_wads3d_cadc}(b) and (d), applying snow filtering yields an average of 6.72\% AP improvement on WADS-3D but results in negligible gains for CADC. This suggests that the greater volume of snow-induced noise in WADS-3D conceals objects and creates clutter; this disrupts the detectors, making filtering essential for performance recovery. In contrast, the lower number of snow points in CADC implies that these points are either less geometrically distinct or their presence does not fundamentally disrupt the feature extraction process of the detection backbone.

These findings provide a nuanced validation of the hypothesis proposed by Vattem et al. \cite{vattem2022rethinking}, who argue that deep learning detectors can inherently learn to distinguish weather artifacts from true objects, making snow filtering obsolete. While our CADC results align with this argument, the performance on WADS-3D reveals a critical threshold. Higher sensor resolution and heavier snowfall creates noise densities that exceed the generalization limits of clear-weather models. Consequently, while end-to-end detectors resist moderate interference, explicit filtering is essential for high-resolution sensors in extreme weather that overwhelms learned feature invariance.

\subsubsection{Architectural Robustness Under Snowfall}

Cross-dataset evaluation reveals a clear link between architectural design and robustness under snowfall. Voxel-based models (SECOND-IoU) consistently achieve the highest performance, outperforming pillar-based (PointPillars), hybrid (PV-RCNN), and point-based (Part-A2) methods. This hierarchy suggests that voxel aggregation within grid cells filters sparse noise, whereas point-based methods remain highly vulnerable to individual point perturbations from falling snow. Furthermore, anchor-based architectures outperform anchor-free frameworks (Part-A2), proving that strong spatial priors stabilize degraded sensor data. These insights provide a practical architectural baseline for model selection and future model development in severe winter conditions.

\subsection{Performance Decomposition \& The Recall Paradox}

\subsubsection{Error Decomposition \& Precision-Recall Analysis}

\begin{figure}[htp] \centering \includegraphics[width=8.5cm]{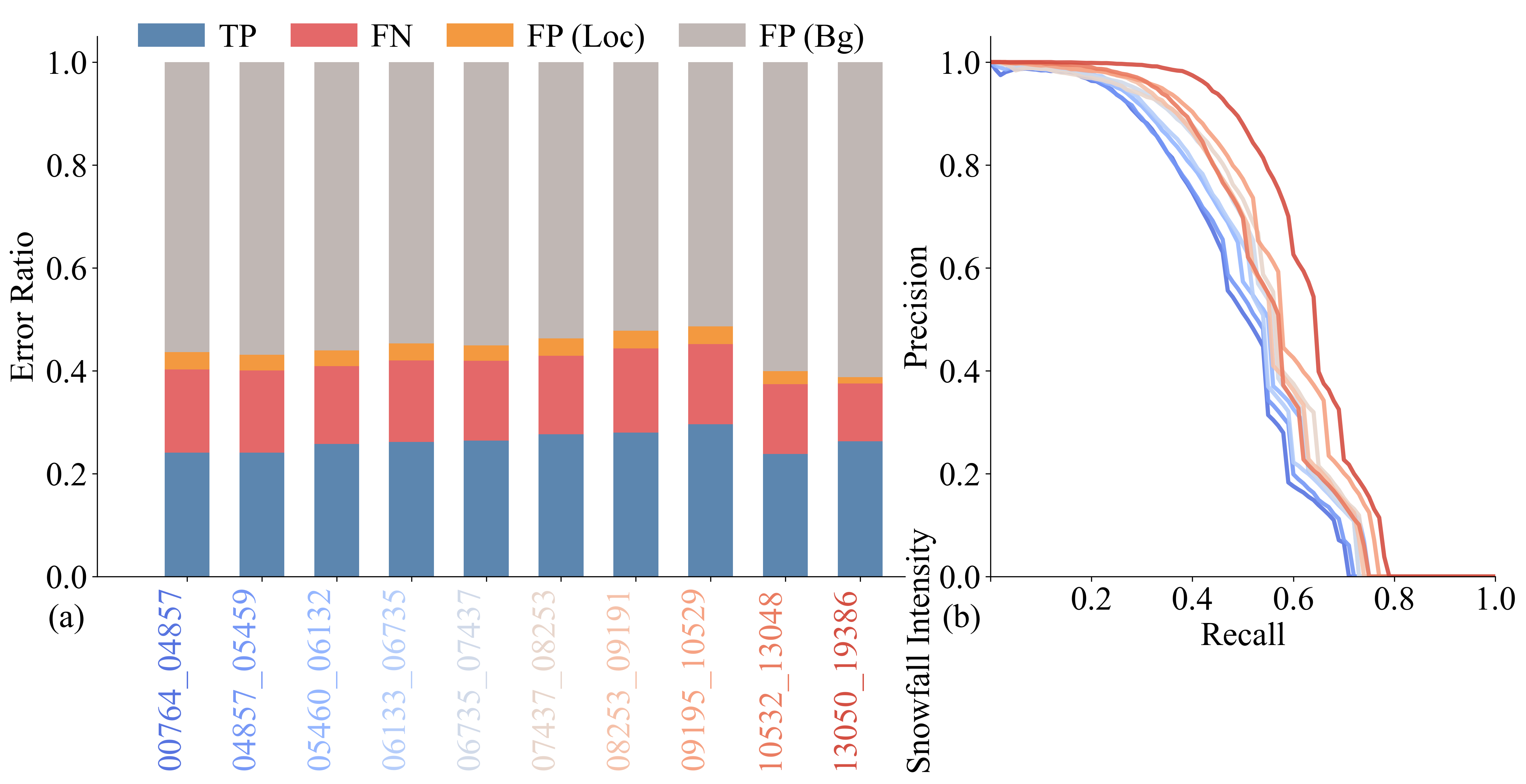} \caption{Normalized error distribution and PR-curve across varying snowfall intensities on WADS-3D.}
\label{fig:error_wads}\vspace{-3mm} \end{figure}

For WADS-3D, as shown in Fig.~\ref{fig:error_wads}(a), a distinct trend emerges as snowfall intensity increases: the proportion of TP increases while FN decreases, indicating a recall improvement. While environmental clutter causes fluctuations in FPs, dominated by Background errors (FP-Bg) rather than Localization errors (FP-Loc), the PR curves in Fig.~\ref{fig:error_wads}(b) shift upwards and to the right for heavier snow intensities, signifying an overall improvement in AP despite the increased snowfall noise.

\begin{figure}[htp] \centering \includegraphics[width=8.5cm]{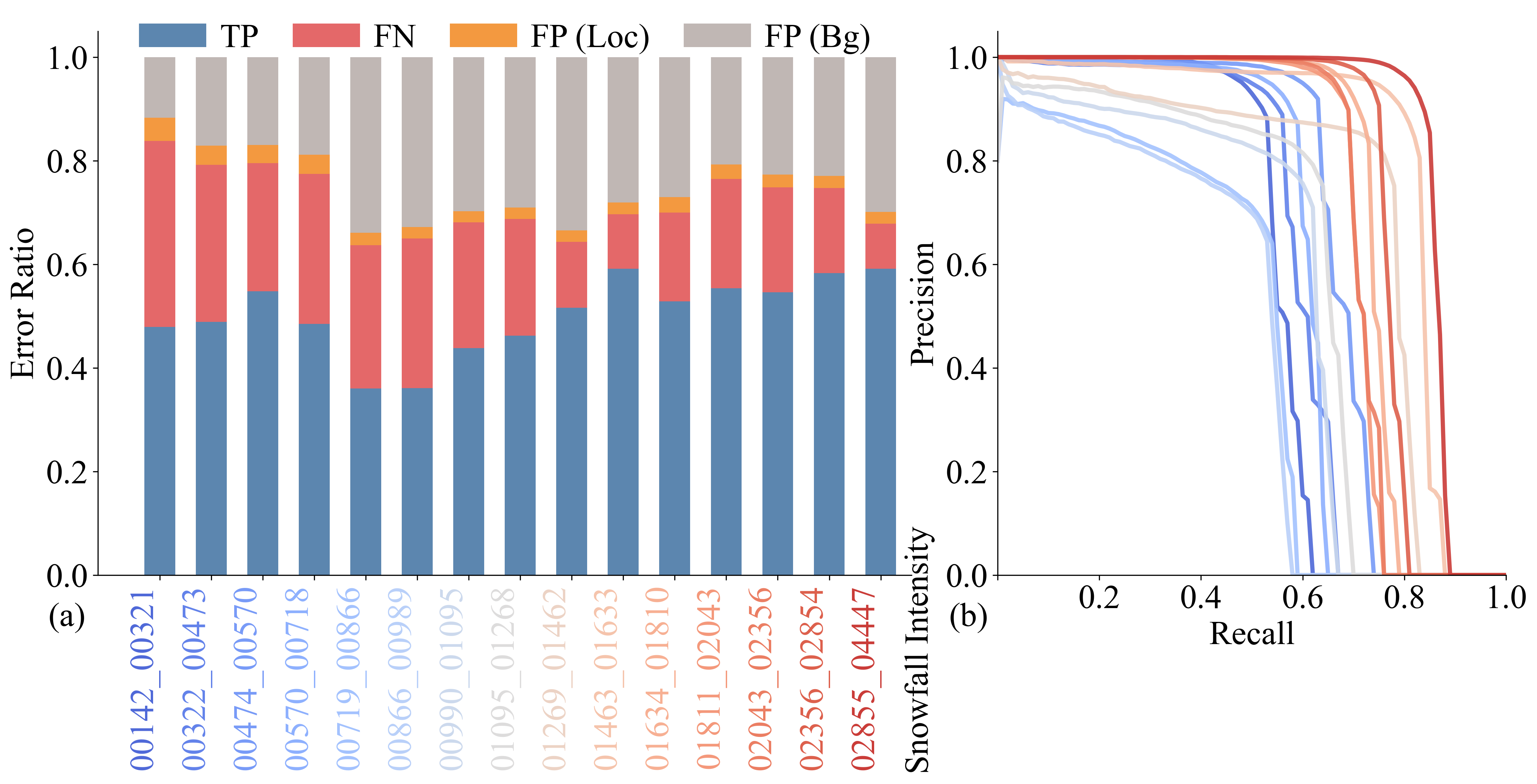} \caption{Normalized error distribution and PR-curve across varying snowfall intensities on CADC. }\label{fig:error_cadc} \end{figure}

This trend is also pronounced in the CADC dataset (Fig.~\ref{fig:error_cadc}). Compared to WADS-3D, CADC exhibits sharper gradients in error distribution; the reduction in FNs is significant and follows a clear overall downward trend across increasing snowfall levels. Consequently, the CADC PR curves (Fig.~\ref{fig:error_cadc}(b)) shifting more outward relative to WADS-3D, delaying precision drop-off and maintaining high precision especially in the high recall regime. These combined results confirm that increased recall driven by increased TP and reduced FN significantly outweighs the influence from increased background clutter (FP), thereby driving the net increase in AP.

\subsubsection{The Physical Paradox} 

These relationships present a physical contradiction. Inherently, atmospheric noise like snow should attenuate LiDAR signals and occlude objects, which typically increases FNs and reduces TPs. The fact that objects appear easier to detect in heavy snow, despite the sensor degradation suggests that the improvement is not driven by sensor physics, but rather by the characteristics of the objects themselves within the heavy snow frames.

\subsubsection{Transition to Meta-Analysis} 

To resolve this paradox, we move beyond aggregate metrics to a fine-grained analysis. The error matrix and PR curves in Fig.~\ref{fig:error_wads} and Fig.~\ref{fig:error_cadc} reveal that the AP improvement is driven almost exclusively by improved recall, specifically due to the recovery of objects in heavy snow. Consequently, we consolidate TPs and FNs into a binary classification framework to perform a meta-analysis on recall. This approach allows us to isolate physical object properties ( e.g. distance, point count, and size) to determine if the heavy snow subset contains objects that are easier to detect, thereby inflating performance.

\subsection{Meta-analysis on Recall}

\subsubsection{Analysis of Recall Determinants on WADS-3D} 

The meta-analysis of detection performance reveals that object point count and object distance are the governing factors for recall across all five evaluated architectures: they exhibit the strongest correlations (Kendall’s $\tau \approx$ 0.3 and $-0.3$, respectively) and highest mutual information scores, indicating that sparsity induced by range is the primary failure mode. SHAP feature importance further corroborates this, attributing $\approx$ 40-50\% of the model's decision-making weight to these two predictors, with object dimensions (length and height) serving as secondary discriminators. Notably, snowfall intensity display negligible feature importance ($<$0.03) and weak correlation across all models. This suggests that recall in WADS-3D is driven primarily by object attributes rather than by the interference of weather noise or the environment cluster.

\begin{figure*}[t] 
  \centering
  \includegraphics[width=0.99\textwidth]{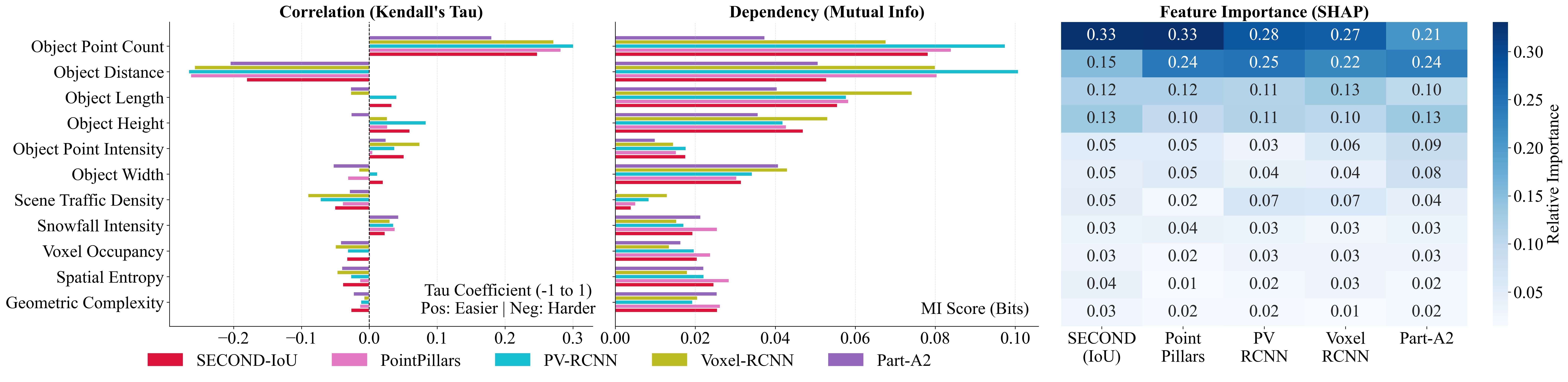} \caption{Kendall's Tau correlation, Dependency (MI), and Feature importance (SHAP) for recall analysis on WADS-3D.} \label{fig:meta_analysis_recall_wads3d}\vspace{-3mm} 
\end{figure*}

\subsubsection{Analysis of Recall Determinants on CADC} 

Consistent with the strong dependence on point sparsity observed in WADS-3D, object point count remains the primary determinant of recall in CADC (SHAP $\approx$ 0.55). However, unlike WADS-3D, detection in CADC exhibits a pronounced dependence on radiometric features; mean point intensity emerges as a secondary discriminator, showing a strong positive correlation (Kendall’s $\tau \approx$ 0.3) and substantial feature importance (SHAP $\approx$ 0.18) across all architectures. This suggests that the detectors refined on CADC also rely on reflectivity cues to separate vehicles from the background. Snowfall intensity continues to play a negligible role. While object dimensions were influential in WADS-3D, they show negligible impact here, likely because larger vehicle dimensions do not translate to increased point returns under the given sensor resolution.

\begin{figure*}[t] 
  \centering
  \includegraphics[width=0.99\textwidth]{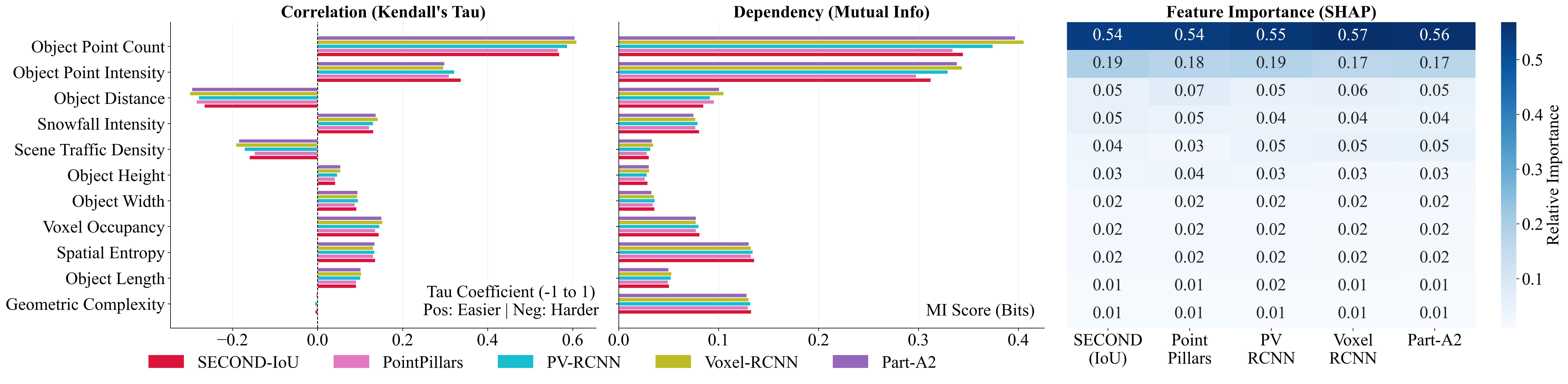} \caption{Kendall's Tau correlation, Dependency (MI), and Feature importance (SHAP) for recall analysis on CADC.} \label{fig:meta_analysis_recall_cadc}\vspace{-3mm}
\end{figure*}

\subsection{Identifying Key Meta-Factors Driving Recall} 

The recall meta-analysis on both datasets reveals that the increasing detection performance is heavily modulated by object-level confounding variables. To quantify this, we investigate the trends of the dominant recall drivers identified in Fig.~\ref{fig:meta_analysis_recall_wads3d} and Fig.~\ref{fig:meta_analysis_recall_cadc}: object point count and ego-distance for WADS-3D, and object point count and mean object intensity for CADC across increasing snow severity levels. We calculate the Pearson correlation ($r$) between the snowfall intensity bins and these key meta-features (Table~\ref{tab:correlation_analysis}).

For WADS-3D, a strong positive correlation exists between snow intensity and average object point count ($r=0.820, p<0.01$), while ego-distance shows no significant trend ($r=-0.282, p=0.43$). This confirms that the counter-intuitive performance is driven by a systematic increase in object point count rather than a reduction in object distance. Similarly, for CADC, both object point count ($r=0.733$) and mean object intensity ($r=0.694$) exhibit significant positive correlations with snowfall intensity. These results provide compelling evidence that objects in higher snow bins are inherently easier to detect due to higher point count and  stronger intensity returns, confounding the evaluation of model robustness under adverse weather.

\begin{table}[htbp]
\caption{Correlation Between Snowfall Intensity and Key Object Meta-Features}
\label{tab:correlation_analysis}
\centering
\begin{tabular}{@{}lllll@{}}
\toprule
\textbf{Dataset} & \textbf{Feature} & \textbf{Pearson $r$} & \textbf{$p$-value} & \textbf{Sig.} \\ \midrule
\textbf{WADS-3D} & Object Point Count      & \textbf{0.820}       & $3.70 \times 10^{-3}$ & ** \\
                 & Object Distance     & $-0.282$             & $0.430$               & NS            \\ \midrule
\textbf{CADC}    & Object Point Count      & \textbf{0.733}       & $1.86 \times 10^{-3}$ & ** \\
                 & Object Mean Intensity   & \textbf{0.694}       & $4.12 \times 10^{-3}$ & ** \\
                \bottomrule
\addlinespace[1ex]
\multicolumn{5}{l}{\small Bold values indicate statistical significance ($p < 0.05$).} \\
\multicolumn{5}{l}{\small Sig. levels: ** ($p < 0.01$), * ($p < 0.05$), NS (not significant).}
\end{tabular}
\vspace{-3mm}
\end{table}

\subsection{Discussion and Hypotheses} While Object Point Count is the primary statistical driver of performance gains, the underlying physical mechanisms remain uncharacterized. We propose three hypotheses for why snow-car interactions might favor detection:

\begin{itemize}

\item \textit{Hypothesis 1: Snow as a Scene Smoother.} This hypothesis suggests that snow acts as a physical filter or smoother for the environment. By coating complex background geometries (e.g., vegetation, irregular road edges), snow may dampen the environmental noise that interferes with car detection. In this scenario, the car's signal becomes more distinct against a simplified background.

\item \textit{Hypothesis 2: Snow as a Geometric Enhancer (Point Count Inflation).} This hypothesis directly addresses the observed increase in Point Count. Due to lidar measurement uncertainty, it is often impossible to differentiate between snowflake and object returns if they are separated by a distance on the order of the lidar's measurement error. As the snowfall rate increases, the density of snowfall points within the car's volume increases. These points may be incorrectly integrated into the car’s bounding box, artificially enhancing its geometry, particularly for smaller objects that would otherwise fall below the detector's threshold.

\item \textit{Hypothesis 3: Surface Reflectance Transformation.} This hypothesis focuses on the optical properties of the car’s surface. We suspect that snow accumulation transforms a car’s surface from a specular (mirror-like) surface to a near-diffuse/Lambertian surface. In clear conditions, lidar pulses could bounce off smooth car paint at high angles of incidence and fail to return to the sensor. A coating of snow scatters light more uniformly, allowing for (1) a higher number of points returned from steep angles and (2) higher intensity returns.

\end{itemize}

To validate these physical hypotheses, we leverage empirical findings from our controlled study\cite{yang2026geometric}, which systematically analyzed the geometric and radiometric characteristics of snow-covered and clean vehicles across four distinct LiDAR wavelengths and resolutions. This analysis confirms that snow accumulation alters object signatures through two primary mechanisms: (1) geometric inflation, where snow accumulation physically increases the vehicle's effective volume, yielding a higher point count, and (2) a radiometric shift, where the transition from the specular-like reflectance of automotive paint to the high-albedo, diffuse-like reflectance of snow enhances return intensity, particularly at high incidence angles and long ranges. These empirical findings confirm that physical snow accumulation constructively alters the vehicle's geometric and radiometric profile.

%% file: sections/conclusion.tex
\subsection{Conclusion}

This work investigates the counter-intuitive improvement of 3D object detection performance under heavier snowfall. Through ablation studies, we confirm that falling snow acts as noise; however, its negative impact is outweighed by a dominant, concurrent factor: significantly increased object point count. Our analysis reveals that heavier snow scenarios correlate with denser object representations, likely driven by surface accumulation and geometric expansion, drastically reduce False Negatives across both WADS-3D and CADC datasets. These findings demonstrate that evaluating perception systems solely via environmental bins, without accounting for object-level shifts, leads to incomplete robustness profiles. We conclude that accurate safety validation requires disentangling precipitation noise from correlated geometric changes, and we suggest future benchmarks to explicitly model these surface-level accumulation effects.

\subsection{Future Work}

Moving forward, we plan to extend this research in three key directions. First, we will evaluate whether physical snow accumulation yields similar geometric enhancements for smaller, non-rigid targets (e.g., pedestrians and postboxes) or if atmospheric attenuation dominates for smaller cross-sections, necessitating targeted data collection. Second, we will investigate practical mitigation strategies, such as dynamic point-density normalization and camera-LiDAR sensor fusion, to counteract the geometric distortions introduced by snow accumulation. Finally, we aim to develop high-fidelity simulation frameworks that model physical snow accumulation on scene geometry rather than relying solely on airborne noise.